\documentclass[10pt, a4paper]{article}
\usepackage{lrec}
\usepackage{graphicx}
\usepackage{tabularx}
\usepackage{soul}

\usepackage{epstopdf}
\usepackage[latin1]{inputenc}

\usepackage{hyperref}
\usepackage{xstring}

\usepackage{amssymb}
\usepackage{pbox}
\usepackage{tikz}
\usetikzlibrary{matrix}
\usepackage{qtree}
\usepackage{tikz-qtree}
\usepackage{stmaryrd}
\usepackage{xfrac}

\title{A Formal Analysis of Multimodal Referring Strategies Under Common Ground}

\name{Nikhil Krishnaswamy and James Pustejovsky}

\address{Brandeis University Department of Computer Science \\
         Walthan, MA, USA \\
         \{nkrishna,jamesp\}@brandeis.edu}

\abstract{
In this paper, we present an analysis of computationally generated mixed-modality definite referring expressions using combinations of gesture and linguistic descriptions.  In doing so, we expose some striking formal semantic properties of the interactions between gesture and language, conditioned on the introduction of content into the {\it common ground} between the (computational) speaker and (human) viewer, and demonstrate how these formal features can contribute to training better models to predict viewer judgment of referring expressions, and potentially to the generation of more natural and informative referring expressions.\\ \newline  }

\begin{document}

\maketitleabstract

\section{Introduction}

Multimodality has been a topic of study in computational linguistics and natural language processing since at least the mid-1990s \cite{johnston1997unification}, but has seen increased interest from the CL/NLP communities in recent years.  This has been due to a number of factors, including the increase in processing power; the availability of large datasets of text, images, and video; the rise of depth sensors (e.g., Microsoft Kinect$\textregistered$); and the availability of GPUs for deep model training.  This has resulted in a number of new datasets and approaches to cross-modal linking \cite{yatskar2016situation,goyal2017something}, shared tasks \cite{barrault-etal-2018-findings}, and grounding tasks \cite{beinborn-etal-2018-multimodal,zhou-etal-2018-visual}.

The most common modalities under study in the CL/NLP communities are text, audio/speech, and images/video, but ``modality" can in principle refer to any channel of information.  Therefore, multi-channel transmission of information can be separated by channel into the particular information transmitted by each modality (i.e., objects depicted in images with their descriptions in text, or spoken demonstratives with aligned deixis via a gesture).  Such disjunct mechanisms allow us to package, quantify, measure, and order our experiences, creating rich conceptual reifications and semantic differentiations.  By examining the nature of these differentiations, we can study the conceptual expressiveness of these systems \cite{pustejovsky2018actions}.

Demonstrating such knowledge is needed to ensure a shared understanding between interlocutors, and when one such interlocutor is a computer whose multichannel expressions are quantitatively defined, this allows us to measure certain aspects of the {\it computational common ground} created by the computer's representation of information it has shared with its interlocutors, including humans.

When two agents are co-situated and attending to the same situation ({\it co-attending}), it is the introduction of such information into the discourse that creates the ``shared situated reference'' \cite{pustejovsky2017creating} between them, and the introduction of particular information into the common ground may be more or less informative depending not only on the prior contents of the common ground but also the {\it modality through which the new information is introduced}.  The task is then to assess this, either quantitatively or formally.

In this paper, we present an analysis of the common ground structures presented in a dataset of {\it Embodied Multimodal Referring Expressions} (EMRE) \cite{krishnaswamy2019generating}.  These are references to definite objects performed by an avatar in a simulated world using gesture, language, or both.  The appropriateness of each referring technique depicted was then evaluated by annotators on Amazon Mechanical Turk.  The virtual environment allows saving a number of quantitative and qualitative parameter values for each depicted referring technique, allowing further analysis, including for our purposes here, of the introduction of elements between the avatar and the annotators (as proxy for the human interlocutors), and the subsequent update to the common ground caused by each new element.  We analyze both the formal parameters of the common ground updates, and their quantitative effects on annotator preferences for referring techniques within the data.

\section{Related Work}

Referring expressions of course pervade natural language dialogues and are a prominent subject of study in natural language processing \cite{krahmer2012computational}.  Dale and Reiter \shortcite{dale1995computational} identify a successful referring expression as one that identifies the intended target to the hearer without introducing false implicatures a la Grice \shortcite{grice1975logic}.  Paraboni et al. \shortcite{paraboni2007generating} discuss generating referring expressions in hierarchically structured domains, and explore the hypothesis that reducing search for the identifying referent with a referring expression can be improved by including logically redundant information, such as denoting the same content using different methods.  Thus a successful referring expression a la Dale and Reiter may not necessarily be quantitatively optimal as long as it is sufficiently Gricean.

Current approaches to referring expressions include neural approaches with high-dimensional word embeddings \cite{ferreira2018neuralreg} and spatial expression generation in human-robot interaction \cite{wallbridge2019generating}---including grounding referring expressions in an environment using visual features and attributives \cite{shridhar2018interactive,cohen2019grounding,magassouba2019multimodal}.

Studies in the interaction between language and gesture also have a long history in computational linguistics \cite{claassen1992generating,bortfeld1997use,van2001generating,krahmer2003new,funakoshi2004generating,viethen2008use}.  Despite this, there has been comparatively little research from the community into the ways that multiple modalities interact during {\it real-time} communication and how to replicate such structures computationally.  Most work in this area originated in the psychology and cognitive science communities, and has been explored in related communities such as robotics \cite{petit2012coordinating,matuszek2014learning,whitney2016interpreting,kasenberg2019generating}, but has direct relevance to computational language understanding and generation.

McNeill \shortcite{mcneill2000language} argues that thought is multimodal, and that the combinatorics of gesture do not correspond to the syntagmatic values that emerge from the combinatorics of speech. Quek et al. \shortcite{quek2002multimodal}, holds that speech and gesture are coexpressive and processed partially independently, and therefore complement each other. Thus, if interlocutors agree that the meaning of a gesture in a description and the meaning of accompanying speech share the same referent, this must be tested to see if 1) the gesture and speech align, and 2) they share the same denotative content.  Thus rather than by abstract combinatoric analysis, the appropriateness of the referencing operation must be tested within a shared common ground.  This is where we feel that both formal and statistical analysis can be used together to establish computational principles for combining multimodal streams while maintaining maximum interpretability.  First we will describe the dataset we examined, then outline the formal principles of computational common ground, and finally present the methodology and results of our analysis.

\section{Embodied Multimodal Referring Expressions}
\label{sec:emre}

\begin{figure}[h!]
    \centering
    \includegraphics[width=.45\textwidth]{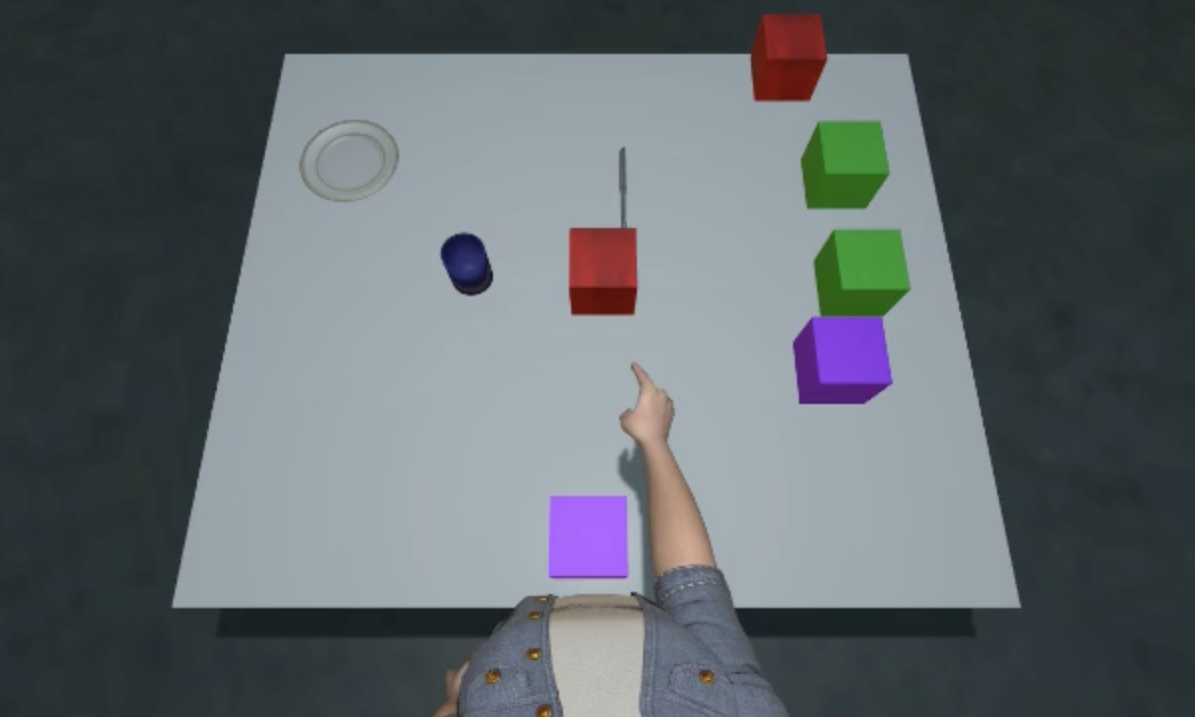}
    \caption{Frame from sample EMRE video, with accompanying utterance ``that red block in front of the knife.''}
    \label{fig:sample}
\end{figure}

Previously, we gathered a dataset of what we called {\it Embodied Multimodal Referring Expressions} (EMRE): that is, visualizations of an agent (here a virtual avatar in a simulated environment) referring to definite objects in her world using various means, including gesture, language, or a multimodal mixture (``ensemble'').  The dataset consists of videos of the avatar using various techniques to refer to a given object in a given configuration in her virtual world, along with associated parameters used in the generation of each video.  The details of the dataset generation process are given in Krishnaswamy and Pustejovsky \shortcite{krishnaswamy2019generating}.\footnote{The dataset itself is available at \hyperlink{https://github.com/VoxML/public-data/tree/master/EMRE}{https://github.com/VoxML/public-data/tree/master/EMRE}}.  In brief, annotators (eight per each video) were presented with a scene depicting objects on a table, given a target object, and then asked to rank each of the depicted methods with which the avatar in the scene then referred to the target object (see Fig.~\ref{fig:sample}).  The avatar used one of the three available modal options (gesture, language, or ensemble), with variants in the language used, to distinguish the target object with regard to its distinct properties or relations to other objects in the scene.  Annotators were asked to rank, on a scale of 1 (least) to 5 (most), the ``naturalness'' of the referring techniques presented relative to the indicated target object, as the goal was to gather data with which to generate multimodal referring expressions in real-time that are appropriate, salient, and natural in context.  Annotation results, links to video, and parameters of each scene depicted are stored in a SQL database.  Stored parameters include some specific to the target object, such as its identity or the distance from it to the simulated agent; some specific to the referring expression, such as modality, utterance used, and relational descriptors in the utterance; and some global to the scene, such as object raw coordinates or total relation set present in the simulation.

\section{Computational Common Ground}

The theory of common ground has a rich and diverse literature concerning what is shared or presupposed in human communication \cite{clark_grounding_1991,stalnaker2002common,asher1998common,tomasello2007shared}. Adopting and extending the model in Pustejovsky \shortcite{pustejovsky2018actions}, given a context of a co-situated interaction, the common ground is a state monad, the components of which are: {\bf A}, the agents engaged in communication; {\bf B}, the shared belief space; {\bf P}, the objects and relations jointly perceived in the environment; and $\mathcal{E}$, the embedding space occupied by the agents.  In the scenario under analysis here, we can specify {\bf A} as \{$\alpha_a$ (the avatar), $\alpha_h$ (the human observer)\}, {\bf B} as $\subseteq$ \{{\it beliefs about the existence, affordances, and relative placement of objects, and the interlocutor's knowledge thereof}\} (the set is denoted as $\Delta$); and {\bf P} as $\subseteq$ \{{\sc Table}, {\sc Cup}, {Knife}, {\sc Plate}, {\sc PurpleBlock1}, {\sc PurpleBlock2}, {\sc RedBlock1}, {\sc RedBlock2}, {\sc GreenBlock1}, {\sc GreenBlock2}, {\it locations within} $\mathcal{E}$\}.  Each element may be introduced into the common ground at any time, such as at $t_0$ or subsequently based on an action taken by one of the agents.  For instance, an agent might introduce a new object into the scene, making common the knowledge of its existence.  Or (as happens in the EMRE dataset), one agent may use certain terms in a definite description, making public their knowledge of the meaning of those terms.

\begin{figure}[h!]
\centering
$\begin{tabular}{|c|}
\hline 
$\begin{array}{lcl}
{\footnotesize \mbox{\bf A:}  \alpha_a, \alpha_h \;\;}
{\footnotesize \mbox{\bf B:} \;\; \Delta \;\;}
{\footnotesize \mbox{\bf P:} \;\; t, c, k, pl, p_1, p_2, r_1, r_2, g_1, g_2}\\
\hline \\
\begin{tikzpicture}
$\,$
\tikzset{every tree node/.style={align=center,anchor=north}}
 \Tree [.$\mathcal{GU}_{\alpha_a}$ [.Point$_g$ [.Dir \textbf{d} ] [.Obj \node(wh){$r_1$}; ] ] ]
\begin{scope}[yshift=-2.7in,grow'=up]
\Tree [.$\mathcal{S}_{\alpha_a}$ [.Dem$_O$ \node(j1){\bf that red block}; ]
               [.PP$_{Loc}$ \node(j2){\bf in front of the knife}; ]]
\end{scope}
\begin{scope}[dashed]
\draw (wh)--(j1);
\end{scope}
\end{tikzpicture}
\end{array}$
 \\
\hline
\end{tabular}_{\mathcal{E}}$\\
\begin{flushleft}
$\lambda k_s \otimes k_g  ({\mbox {\bf that}}(x)[\mbox{block}(x) \wedge \mbox{red}(x) \wedge \mbox{in\_front}(x,k,v)] \wedge k_s \otimes k_g (x)]$,   where $v = \alpha_a$
\vspace{-3mm}
\end{flushleft}
\caption{Common-ground structure for ``that red block in front of the knife'' (cf. Fig.~\ref{fig:sample}). The semantics of the RE includes a {\it continuation} (in the abstract representation sense in computer science, cf. Van Eijck and Unger \shortcite{van2010computational}) for each modality, $k_s$ and $k_g$, which will apply over the object in subsequent moves in the dialogue.}
    \label{fig:cgs-deixis}
\end{figure}


Given the common ground, a communicative act $C_{\alpha}$, performed by agent, $\alpha$, is a tuple of expressions from the modalities available to $\alpha$, involved in conveying information to another agent.  Here, we restrict this to the modalities of a linguistic utterance, $\mathcal{S}$, and a gesture, $\mathcal{G}$. 
Thus there are three possible configurations in performing $C$:
\begin{enumerate}
\vspace{-2mm}
\item $C_\alpha = (\mathcal{G})$
\vspace{-2mm}
\item $C_\alpha = (\mathcal{S})$
\vspace{-2mm}
\item $C_\alpha = (\mathcal{S},\mathcal{G})$
\end{enumerate}

\noindent In the case of co-gestural speech, $(\mathcal{S},\mathcal{G})$, we assume an aligned language-gesture syntactic structure, for which we provide a continuized semantic interpretation. Both of these are contained in the common ground state monad (see Fig.~\ref{fig:cgs-deixis}).

In co-gestural speech, the modal channels can be {\it aligned} or {\it unaligned}.   Each input updates the common ground  and each update to the common ground may change the probability of a subsequent communicative act being more or less salient, based on the content that {\it it} introduces into the common ground.  Thus we propose that the formal characteristics of common ground updates serve as predictors of the naturalness of a referring expression, based on the saliency of the content communicated through the update.

Common ground updates execute modal operations over the belief space {\bf B} such that each element of $\Delta$ is introduced via a {\it public announcement logic} (PAL) formula or an analogous formula denoting what the agents see or perceive \cite{plaza2007logics,van2007dynamic,van2011logical}.  To avoid confusion between the two, we use the standard syntax of Plaza's public announcement logic with the following exceptions: we will use $\mathcal{K}_\alpha \varphi$ to denote ``$\alpha$ \mbox{\it knows} $\varphi$'', 
 $\mathcal{L}_\alpha \varphi$ to denote ``$\alpha$ \mbox{\it   believes} $\varphi$'', and $\mathcal{P}_\alpha \varphi$ to denote ``$\alpha$ \mbox{\it perceives} $\varphi$''. These are employed in place of a generic doxastic/epistemic update $[\alpha]\varphi$ (``agent $\alpha$ knows/believes $\phi$"), so that an utterance like ``You see it," as in Fig.~\ref{fig:cgs1}, serves to express the update $[\mathcal{K}_{\alpha_h} \mathcal{P}_{\alpha_a} b!]\mathcal{K}_{\alpha_h} \mathcal{P}_{\alpha_a} b$, glossed as ``$\alpha_h$ publicly announces (indicated by the bang, $!$) that $\alpha_h$ knows  $\alpha_a$ perceives $b$."

\begin{figure}[h!]
\centering
$\begin{tabular}{|c|}
\hline 
$\begin{array}{lcl}
{\footnotesize \mbox{\bf A:} \;\; \alpha_a, \alpha_h \;\;}
{\footnotesize \mbox{\bf B:} \;\; \Delta \;\;  }
{\footnotesize \mbox{\bf P:} \;\; b \;\; }\\
\hline \\
 \mathcal{S}_{\alpha_h} = \mbox{``}\mbox{You}_{\alpha_a} \; \mbox{see} \; \mbox{it}_{b}\mbox{"}
 \end{array}$
 \\
\hline
\end{tabular}_{\mathcal{E}}$
\caption{Common-ground structure for  ``You see it.''}
    \label{fig:cgs1}
\end{figure}

\begin{figure}[h!]
    \centering
    \includegraphics[width=.23\textwidth]{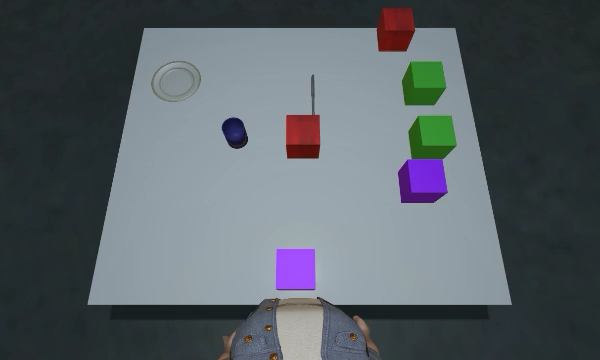}
    \includegraphics[width=.23\textwidth]{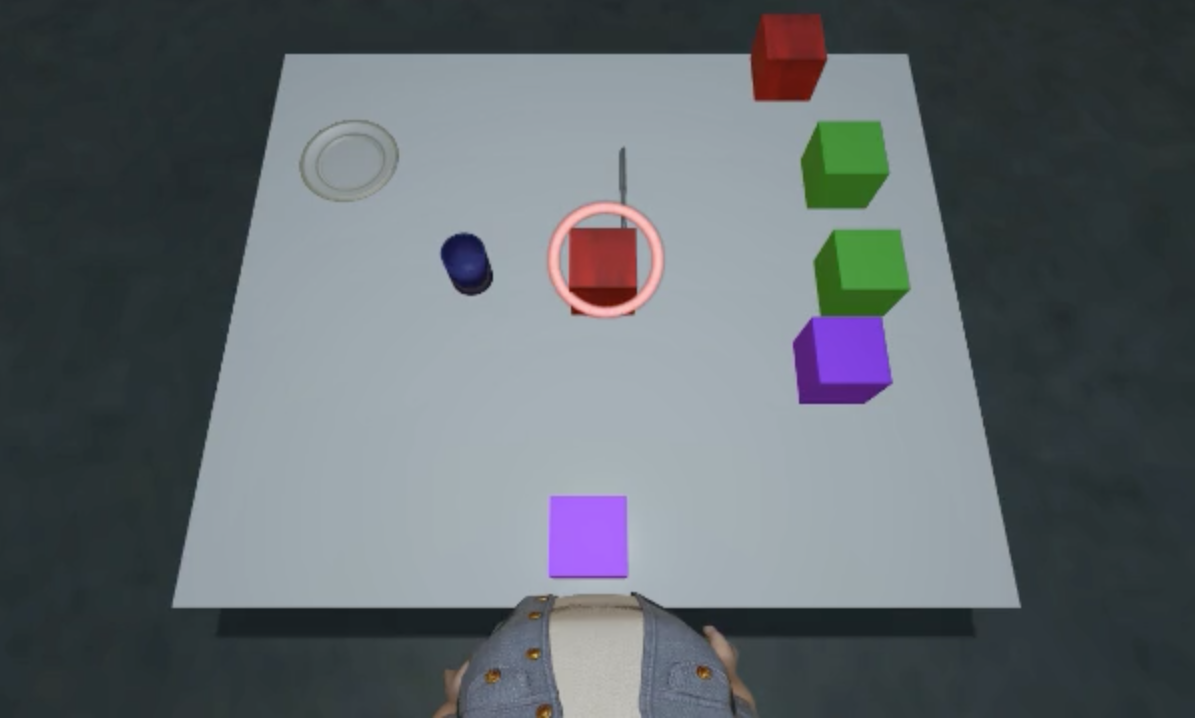}
    \caption{Additional frames accompanying common ground updates.}
    \label{fig:sample2}
\end{figure}

The types of update that we will examine as pertaining to referring expressions (REs) in the EMRE dataset are:
\begin{enumerate}
    \item At $t_0$, the beginning of each video, the scene is presented.  All objects displayed populate {\bf P}, the elements of the jointly-perceived environment. $\forall b$ ($b \in$ {\bf P} $\rightarrow \mathcal{K}_{\alpha_h} \mathcal{P}_{\alpha_a} b \wedge \mathcal{K}_{\alpha_a} \mathcal{P}_{\alpha_h}$).  This is shown in Fig.~\ref{fig:sample2} (L). This is derived by performing transitive closure of perception over the agents who are co-situated in the perceptual environment:  
     $[(\mathcal{P}_{\alpha_h} \cup \mathcal{P}_{\alpha_a})^*] \phi$.
    \item At $t_1$, a circle is drawn around one particular object ($b$), raising it to the status of target object.  The human observer $\alpha_h$ now knows that $b$ is the target (but does not necessarily know that the avatar $\alpha_a$ knows this as well).  $\mathcal{K}_{\alpha_h} target(b) \wedge \neg \square \mathcal{K}_{\alpha_h} \mathcal{K}_{\alpha_a} target(b)$.  This is shown in Fig.~\ref{fig:sample2} (R).
    \item At $t_2$ (shown above in Fig.~\ref{fig:sample}):
    \begin{enumerate}
        \item The avatar points to $b$.  This demonstrates the avatar can point, and knows that $b$ is the target.  [$C_{\alpha_a} = Point_g \rightarrow Dir \; b$!]$\mathcal{K}_{\alpha_h} \mathcal{K}_{\alpha_a} (Point_g \wedge target(b))$.
        \item The avatar describes $b$ using some combination of $b$'s attributes (here, color), and relations to other objects.  This demonstrates that $\alpha_a$ knows the meaning of the terms she uses ($\llbracket u \rrbracket$ being the {\it interpretation} of some utterance $u$) under a model $\mathcal{M}$ and a common ground $cg$, and also situates some of those terms (e.g., spatial relations) relative to her frame of reference.   [$C_{\alpha_a} = \mathcal{S}$!]$\forall u(u \in \mathcal{S}\rightarrow\mathcal{K}_{\alpha_h} \mathcal{K}_{\alpha_a} \llbracket u \rrbracket_{\mathcal{M},cg})$.
    \end{enumerate}
\end{enumerate}

Time intervals in the video data are all constant, allowing us to maintain consistent timesteps in analysis: the initial presentation ($t_0$) is shown for 1.5 seconds, the target circle is drawn and held for 1.5 seconds ($t_1$), and .5 seconds later, at $t_2$, the agent indicates the target object, through gesture, language, or both.

\section{Analysis Methodology}
\label{sec:methodology}

The EMRE dataset distribution contains an analysis script to evaluate the probability of a given annotator judgment over arbitrary sets of parameters in the scene.  Parameters are presented in the form of SQL conditions to filter the results over, allowing the extraction of conditional probabilities over any parameters recoverable from the data using standard SQL syntax.  Krishnaswamy and Pustejovsky \shortcite{krishnaswamy2019generating} presents a basic statistical analysis of annotator judgments over parameters directly stored in the data.  A full description of all parameters examined is contained therein.  These data showed a clear preference for referring expressions using the gesture+language ``ensemble" modality, and preference for longer descriptive strings, providing coarse-grained parameters over which to train a deployable multimodal referring expression generation model.  However, we believe that examining the formal properties of the referring expressions shown in the data provides further discriminative features for better generation, as indicated by optimized saliency, naturalness, and informativity of the generated expression to human interlocutors.  We extract formal and propositional values as features from the data based on the information each feature introduces into the common ground.  If it is inferable from the content denoted by either $\mathcal{G}$ or $\mathcal{S}$ modality in the referring expression that an agent $\alpha$ either knows or perceives some propositional content $p$ relative to the belief space {\bf B} or the jointly perceived entities in {\bf P}, this prompts an update to the common ground, and therefore new features for possible examination.  Some examples include the following:

\begin{itemize}
    \item Through use of a spatial term $T$, $\alpha$ introduces that interpretation of spatial term $T$ into the common ground: $C_\alpha$ = ($\mathcal{S} \mid T_{sp} \in \mathcal{S}$) $\rightarrow$ $\mathcal{K}_\alpha \llbracket T_{sp} \rrbracket_{\mathcal{M}}$

    \item Through use of an attributive term $T$, $\alpha$ introduces that interpretation of attributive term $T$ into the common ground: $C_\alpha$ = ($\mathcal{S} \mid T_{att} \in \mathcal{S}$) $\rightarrow$ $\mathcal{K}_\alpha \llbracket T_{att} \rrbracket_{\mathcal{M}}$

    \item By referencing $b$ in a description string, $\alpha$ introduces that she perceives $b$ into the common ground (this is different from $b$ itself, which is already in {\bf P}, the set of jointly-perceived objects; this explicitly encodes the {\it knowledge} that $\mathcal{P}_\alpha b$): $C_\alpha$ = ($\mathcal{S} \mid b_s \in \mathcal{S}$) $\rightarrow$ $\mathcal{P}_\alpha b$

    \item By differentiating two similarly colored objects (e.g., by use of ``other"), $\alpha$ introduces that she knows the same attribute predicates over $b_1$ and $b_2$ but that $b_1$ and $b_2$ are distinct: $C_\alpha$ = ($\mathcal{S} \mid [``other",b_{1_s},b_{2_s}] \in \mathcal{S} \wedge b_{1_s} = b_{2_s}$) $\rightarrow$ $\mathcal{K}_\alpha \llbracket Att(b_1 \wedge b_2) \rrbracket_{\mathcal{M}} \wedge \mathcal{K}_\alpha b_1 \neq b_2$

    \item By distinguishing demonstratives in an ensemble RE, $\alpha$ introduces that she is meaningfully distinguishing between ``near" and ``far" regions of $sfc$, the table surface: $C_\alpha$ = ($\mathcal{S}$, $\mathcal{G} \mid \mathcal{G} = Point_g \wedge ``this" \in \mathcal{S}$) $\rightarrow$ $\mathcal{K}_\alpha \llbracket near(sfc) \rrbracket \neq \llbracket far(sfc) \rrbracket_{\mathcal{M}}$
\end{itemize}

The visualizations in the EMRE dataset are produced using the VoxSim event/agent simulation platform \cite{krishnaswamy2016multimodal,krishnaswamy2016voxsim}, which employs the VoxML modeling language, enabling object and event visualization semantics \cite{pustejovsky2016LREC}.  Because a simulator is an extension of a model checker \cite{pustejovsky2014generating}, a simulation can be evaluated formally.  Because a simulator requires numerical parameter values to run \cite{davis2016scope}, it can be evaluated quantitatively.  Values extracted from the simulator and collated in the dataset may be either real numbers or vector values (e.g., distance values or coordinates) or symbolic (e.g., object labels or qualitative attributes).  Thus, we can conduct ablation tests on the effects of formal, symbolic, and quantitative features on the predictive model trained over data extracted from a simulation.

Each of these features and others can be extracted from a common ground structure of the kind shown in Figs.~\ref{fig:cgs-deixis} or~\ref{fig:cgs1}.  In addition, they can be linked with each other and other features by virtue of the linkages established in the common ground structure (e.g., the link between $Dem_O \rightarrow$ {\bf that red block} and $Point_g \rightarrow Obj \rightarrow r_1$ in Fig.~\ref{fig:cgs-deixis}), effectively allowing us to search the data for sets of related parameters that predict given annotator ratings of a referring technique, formally reanalyze them in terms of computational common ground, and use segments of common ground structures as input features into a prediction algorithm.

\subsection{Model Architecture}

Here, the extracted data and quantitative features are those described in Section~\ref{sec:emre}  The formal common ground features are those as described above.  What we are trying to predict, then, is the likelihood of an annotator evaluating a referring expression at a given naturalness (1-5), with the expectation that the best or most natural REs will come with a consistent set of features that predict a high score.

We feed all features into a multilayer perceptron (MLP) written in Keras with the TensorFlow backend.  Our reasons for choosing this type of architecture is its relative simplicity, and therefore training speed, but also ability to distinguish dependencies between points in linearly-inseparable regions of data \cite{cybenko1989approximation}.  Our architecture consists of three fully-connected hidden layers of 32, 128, and 64, respectively, prior to a $softmax$ output layer.  The layers use $tanh$, ELU, and $tanh$ activation, respectively.  The model uses categorical cross-entropy loss and Adam optimization, and is trained for 1000 epochs with a batch size of 50.  Due to the relatively small size of the sample data, we validate all results using 7-fold cross-validation in order to achieve a more balanced sample across all classes of annotator judgments.  $k = 7$ is chosen here to approximate a leave-one-out cross-validation approach over the 8 annotator judgments on each visualized referring expression.  Because in the EMRE dataset, 8 separate annotators evaluated each RE, the ``most likely" annotator judgment is in fact a probability distribution.  Therefore, we regard a ``correct" prediction by the classifier not as one that returns the exact integer value representing the argmax of all annotator judgment counts, but one that falls within the correct quintile of the distribution over all annotator judgments of that visualized referring expression.

\section{Results and Evaluation}

\subsection{Baseline}

The EMRE dataset already contains quantitative and some qualitative features about scenes, referring expressions generated within them, and annotator judgments thereof.  As a baseline, we used the raw features used in the generation of videos in the EMRE dataset to try and predict the most likely annotator rating of that video.  The baseline features include: 1) the target object; 2) the referring modality: one of gesture, language, or ensemble; 3) the distance from the object to the agent; 4) whether the linguistic description uses a near/far distance distinction; 5) whether that distinction is relative to embedding space of similar objects or the entire world ({\it n/a} if no distance distinction is used.  We also add in 6) the linguistic description used and 7) the individual relational descriptors used, which are represented as 200-dimensional sentence vectors trained using a Skip-Gram model over the entire vocabulary that occurs in the dataset.  In gesture-only referring expressions, where all the avatar does is point to the target object, these are vectors of all 0s.

The top half of Table~\ref{tab:baseline-formal} (see Section~\ref{ssec:formal}) shows baseline results.  The raw features extracted from the EMRE dataset are successful in predicting the correct quintile of annotator judgment of the associated multimodal referring strategy approximately \sfrac{2}{3} of the time.  Interestingly, the addition of sentence embeddings caused the average accuracy to drop about 3.28\%, suggesting that sentence embeddings caused some confusion in the classifier.  Discussion of these results follows in Section~\ref{sec:disc}

\subsection{Formal Features}
\label{ssec:formal}

To keep track of formal features, such as those described in Section~\ref{sec:methodology}, we maintain two lists of the propositional content within the common ground structure available to each agent (i.e., what each agent---here the avatar and the annotator---knows and perceives about the scene and about each other).  Each element in these lists is correlated with features extracted from the EMRE dataset using the provided analysis script (refer to Section~\ref{sec:methodology}) and the value is inserted into a data structure representing a common ground structure of the form shown in Fig.~\ref{fig:cgs-deixis}: consisting of a gesture, the speech string, and links between the constituents of each.

The encoding of formal features is done by creating one-hot vectors representing the state of the belief space {\bf B} ($\Delta$) as it pertains to the agents {\bf A} and jointly perceived content {\bf P}.  That is, propositional content that is formally denoted as $C_\alpha$ = ($\mathcal{S}$, $\mathcal{G} \mid \mathcal{G} = Point_g \wedge ``this" \in \mathcal{S}$) $\rightarrow$ $\mathcal{K}_\alpha \llbracket near(sfc) \rrbracket \neq \llbracket far(sfc) \rrbracket_{\mathcal{M}}$, as above, is treated as a one-hot vector for $\alpha$'s knowledge of the distance distinction of $near(sfc)$ and $far(sfc)$ in $\Delta$, whereas content formally denoted in the form $C_\alpha$ = ($\mathcal{S} \mid [``other",b_{1_s},b_{2_s}] \in \mathcal{S} \wedge b_{1_s} = b_{2_s}$) $\rightarrow$ $\mathcal{K}_\alpha \llbracket Att(b_1 \wedge b_2) \rrbracket_{\mathcal{M}} \wedge \mathcal{K}_\alpha b_1 \neq b_2$ is treated as {\it three} one-hot vectors, one for $\mathcal{K}_\alpha \llbracket Att(b_1) \rrbracket_{\mathcal{M}}$, another for $\mathcal{K}_\alpha \llbracket Att(b_2) \rrbracket_{\mathcal{M}}$, and a third for $\mathcal{K}_\alpha b_1 \neq b_2$.

MLP prediction results of annotator judgment using formal features are shown below.  However, if performance increases when the formal features are added, it could be due to the fact that since they are (under our hypothesis) dependent features, and they reinforce each other, giving stronger prediction results.  Therefore, to demonstrate the effect of formal features, we present ablative results using raw features with formally-derived features, raw features with formally-derived features including sentence embeddings, and formally-derived features only.

We present, as before, the mean and standard deviation of classification accuracy over a 7-fold cross-validated sample. 

\begin{table}[h!]
    \centering
    \begin{tabular}{|l|l|l|}
    \hline
         & {\bf Raw features} & {\bf Raw feat. + SE} \\
         \hline
         {\bf $\mu$ Acc. (1K)} & $0.6757$ & $0.6429$  \\
         \hline
         {\bf $\sigma$ Acc. (1K)} & $0.0230$ & $0.0111$  \\
    \hline
    \end{tabular}
    
    \begin{tabular}{|l|l|l|l|}
    \hline
         & {\bf Raw + } & {\bf Raw + } & {\bf Formal} \\
         & {\bf form.} & {\bf form. + SE} & {\bf only} \\
         \hline
         {\bf $\mu$ Acc. (1K)} & $0.7214$ & $0.6671$ & $0.7471$  \\
         \hline
         {\bf $\sigma$ Acc. (1K)} & $0.0398$ & $0.0243$ & $0.0269$  \\
    \hline
    \end{tabular}
    \caption{Classification accuracy after 1000 epochs using formal features (mean and standard deviation over 7-fold cross-validated sample)}
    \label{tab:baseline-formal}
\end{table}

Features that correlate formally with elements of the common ground structure equivalent to the referring strategy depicted do between 7-11\% better at predicting the category label (annotator judgment) on the referring strategy than raw features alone, or raw features augmented with sentence embeddings.

The above data shows that formal features derived from the common ground structure provide a modest but appreciable improvement in the quality of predicting {\it how well} a referring strategy is likely to be perceived as natural, based on the content it encodes, but tells us little about {\it what} propositional content is likely to produce a natural, salient multimodal referring expression, and {\it how} it should be assembled, which is an important question for generating multimodal REs.

Lascarides and Stone's formal semantics of gesture \cite{lascarides2009formal} separates gestural and speech assignment functions in order to distinguish entities that can satisfy interpretations of referents in speech from entities used to ground references in gesture. It should be pointed out that we are focusing on {\it co-gestural speech} ensembles rather than {\it co-speech gesture} \cite{schlenker2018gesture}. Further, since here we focus only on {\it deictic} gesture rather than depicting gestures, we do not evaluate gesture that conflicts semantically with the speech, but we can draw some inferences analogically regarding the information provided by each. 

\begin{enumerate}
    \item If information provided by gesture is constant between referring expressions for the same object, then the ``best" ensemble RE should be that which maximizes the score of its linguistic component if taken alone.
    \item If information provided by gesture is {\it not} constant between referring expressions for the same object, the ``best" ensemble RE should be that which maximizes the information gain provided by each of the individual modalities.
\end{enumerate}

If (1) is true, then we should expect that features dependent only on the language, including sentence embeddings but not including things like the distance from the agent to the target object, should predict the quality of linguistic-only referring expressions better than the full set of features, including parameters dependent on the gesture and embodiment of the agent, predict the quality of ensemble referring expressions.  If (2) is true, this should not be the case.  Given that the gestural component of a well-formed ensemble referring expression is always deixis, we hypothesize that the gestural information is constant across referring expressions.  To test this, we run different subsets of the total raw and formal feature set (depending on which modality each feature explicitly depends on) through the classifier, over either only the linguistic-only referring expressions from the EMRE dataset, or over only the ensemble referring expressions.  Classifier results are given below.

\begin{table}[h!]
    \centering
    \begin{tabular}{|l|l|l|}
    \hline
         & {\bf Raw features} & {\bf Raw feat. + SE} \\
         \hline
         {\bf $\mu$ Acc. (1K)} & $0.7471$ & $0.6329$  \\
         \hline
         {\bf $\sigma$ Acc. (1K)} & $0.0468$ & $0.0577$  \\
    \hline
    \end{tabular}
    
    \begin{tabular}{|l|l|l|l|}
    \hline
         & {\bf Raw + } & {\bf Raw + } & {\bf Formal} \\
         & {\bf form.} & {\bf form. + SE} & {\bf only} \\
         \hline
         {\bf $\mu$ Acc. (1K)} & $0.7471$ & $0.6443$ & $0.7985$  \\
         \hline
         {\bf $\sigma$ Acc. (1K)} & $0.0213$ & $0.0469$ & $0.0405$  \\
    \hline
    \end{tabular}
    \caption{Classification accuracy after 1000 epochs using formal features and linguistically-dependent features only, over purely linguistic EMRE referring expressions (mean and standard deviation over 7-fold cross-validated sample)}
    \label{tab:ling-only}
\end{table}

\begin{table}[h!]
    \centering
    \begin{tabular}{|l|l|l|}
    \hline
         & {\bf Raw features} & {\bf Raw feat. + SE} \\
         \hline
         {\bf $\mu$ Acc. (1K)} & $0.6014$ & $0.5842$  \\
         \hline
         {\bf $\sigma$ Acc. (1K)} & $0.0537$ & $0.0281$  \\
    \hline
    \end{tabular}
    
    \begin{tabular}{|l|l|l|l|}
    \hline
         & {\bf Raw + } & {\bf Raw + } & {\bf Formal} \\
         & {\bf form.} & {\bf form. + SE} & {\bf only} \\
         \hline
         {\bf $\mu$ Acc. (1K)} & $0.6014$ & $0.5842$ & $0.6171$  \\
         \hline
         {\bf $\sigma$ Acc. (1K)} & $0.0302$ & $0.0840$ & $0.0550$  \\
    \hline
    \end{tabular}
    \caption{Classification accuracy after 1000 epochs using formal features, over only ensemble (multimodal) EMRE referring expressions (mean and standard deviation over 7-fold cross-validated sample)}
    \label{tab:ens-only}
\end{table}

Tables~\ref{tab:ling-only} and~\ref{tab:ens-only} demonstrate that not only do linguistically-dependent features predict the quality of language-only referring expressions better than {\it all} features predict ensemble referring expressions, meaning that the level of information provided by solely deictic gesture is likely to be of roughly constant relevance across the dataset (i.e., directly grounding to a location and object(s) in that location), but that the addition of formal features provide a larger net increase in classifier accuracy over the raw feature baseline for the language only REs than they do for the ensemble REs.

\begin{figure}[h!]
    \centering
    \includegraphics[width=.45\textwidth]{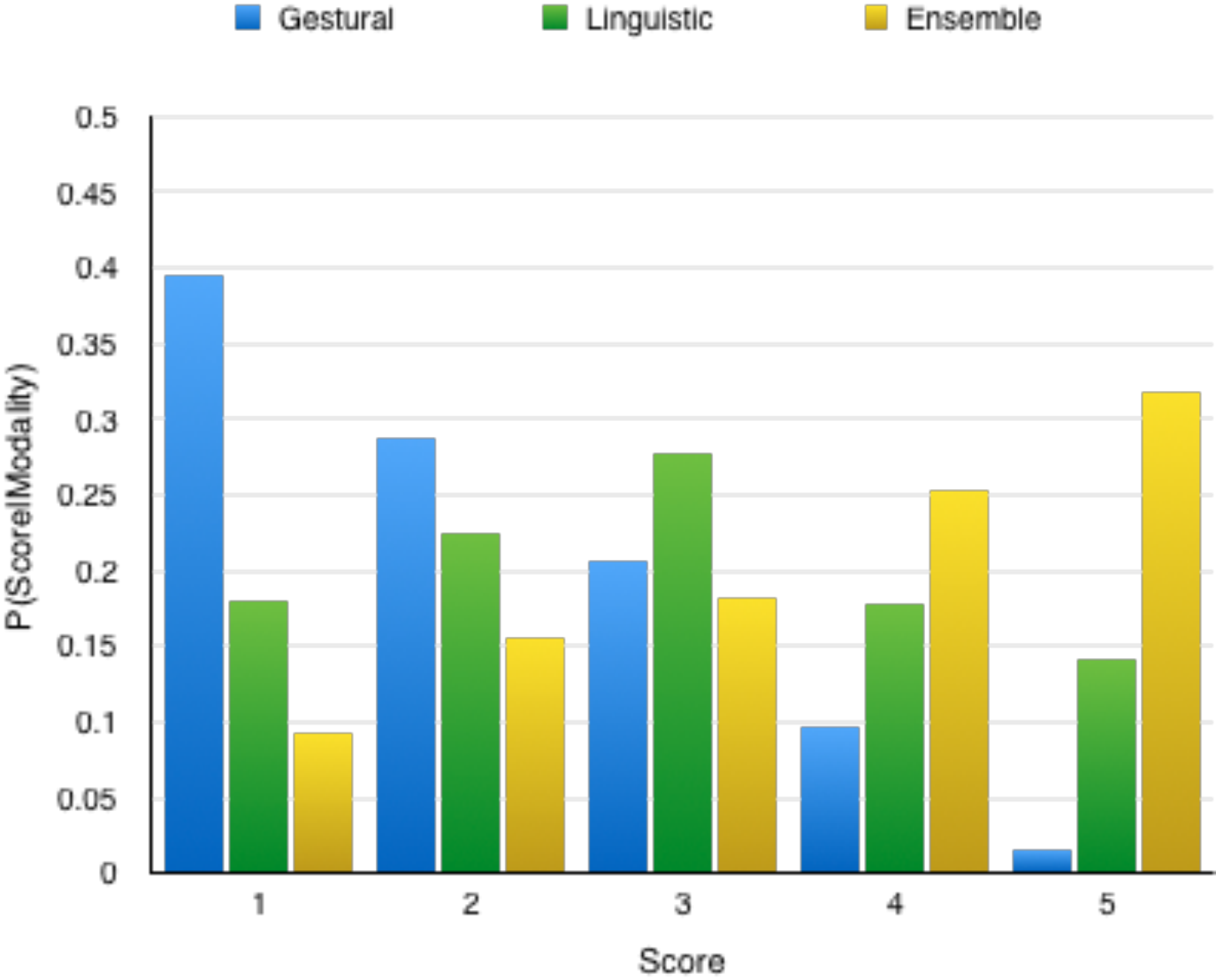}
    \caption{Probability of annotator judgment score given referring expression modality (taken from Krishnaswamy and Pustejovsky (2019))}
    \label{fig:score-modality}
\end{figure}

For linguistic REs, adding formal features to raw features plus sentence embeddings improved accuracy by only about 1\%, but {\it only} using formally-derived features improved accuracy by approximately 5-16\%, depending on if the baseline compared includes sentence embeddings or not.  
For ensemble REs, the addition of formal features made no difference in the average classification accuracy compared to simply raw features (with or without sentence embeddings) and formal features alone resulted in a small ($\sim$1\%) average improvement over the baseline.  From these results we can see that since the referring modality is already a strong predictor of referring expression naturalness and salience according to the dataset (see Fig.~\ref{fig:score-modality}, taken from \cite{krishnaswamy2019generating}), most of the improvement in the multimodal referring expressions compared to low-rated gesture-only referring expressions, where the avatar just wordlessly points to the target object, comes from the information gain associated with an appropriately informative linguistic utterance accompanying the co-speech gesture, which is an important consideration to take into account when generating quality referring expressions, particularly multimodally.

\section{Discussion and Conclusions}
\label{sec:disc}

Our results demonstrate an appreciable increase in the ability of a formal feature set derived from a common ground structure to predict the naturalness and salient quality of a referring expression associated with that common ground structure.  We hypothesize that this is because formal features make the model explainable on a finer-grained level, and that the propositional content extractable from the linguistic utterances used correlates more closely with the quality of the overall referring expressions than less-symbolically defined features like sentence embeddings.

Basic features provide a solid baseline upon which to improve RE classification accuracy \cite{zhang2016introducing}.  Here, however, using sentence embeddings actually seemed to hurt the accuracy.  In the data, the purely linguistic ``the red block in front of the knife" is more likely to be rated as ``average" while ``that red block in front of the knife" is multimodal (accompanied by deictic gesture) and more likely to receive a high rating (see Fig.~\ref{fig:score-modality}).  However, the sentence embeddings for these sentences are very similar, due to an alternation of two words (``the"/``that") that already tend to be similar in distributional semantic space.  ``The" vs. ``that" captures little of the distinction introduced by the ensemble in the common ground.  For this data and task, therefore, this suggests that either simple sentence embeddings are not a very useful feature or should should be trained in a different way, other than a Skip-Gram model.

Meanwhile the formally-defined features extracted from common-ground structures, are much more adept at distinguishing the types of salient information introduced by the referring agent into the common ground and by encoding what type of knowledge is introduced or publicly perceived in the common ground, we are able to quite effectively predict how our annotators would judge the depicted referring expression.

Using the formal features alone usually performs best at this task, likely since the common ground structure is designed specifically to capture the type of information we seek to disambiguate in a multimodal referring expression classification task, compared to raw features that describe either the physical environment or vague contours of the priors that go into the referring expression generation procedure in the EMRE dataset.  Thus we propose that formal common ground structures would be an effective medium through which to interpret and generate multimodal referring expressions and other types of multimodal communicative acts in a co-situated interaction.

\section{Future Work}

The composition of gesture and speech plays an important role in multimodal communication.  The two modalities display complementary strengths at communicating different types of information---it is hard to communicate certain types of spatial configurations solely through language, and deictic gesture may prove more economical; conversely, attributives like color are much more aptly communicated through language.  It is through the combination of the two that successful referring expressions can be generated in co-situated space, and by digging into the data we previously gathered, we have found evidence that while the addition of gesture provides a boost in naturalness and salient quality of a referring expression in co-situated space, the best and most natural REs are those that maximize the salience and naturalness of their linguistic components, even if the linguistic information overlaps with the gestural information (cf. Paraboni et al. \shortcite{paraboni2007generating}).

As such, given that we have trained a prediction model to expose these considerations, the next step is to train a generation model that can be deployed ``live" in a multimodal interaction where the situation encountered at any given time may not cleanly map to a situation from the EMRE dataset.

The process of maximizing the contextually-salient information content provided by the linguistic component of the multimodal referring expression, and by extension by all modalities including iconic gesture and action, could be handled by a composing and constructing expressions with a probabilistic grammar.  

Existing work in multimodal grammars, particularly on gesture and speech (cf. Alahverdzhieva et al.~\shortcite{alahverdzhieva-etal-2012-multimodal}, Alahverdzhieva et al.~\shortcite{alahverdzhieva2017aligning}) often focuses on timing and aligning the gesture and speech components using edge-based constraints to generated a syntax tree of both speech and gesture.  To this we would propose the addition of a continuation-based semantics \cite{krishnaswamy2019multimodal} to capture additional content from common ground structures, such as the formally-derived features that we have shown here can be stronger predictors of gesture-speech ensemble quality, particularly in the domain of referring expressions.

Given demonstrated success in the prediction of multimodal referring expression quality, for which formally-derived features are an asset, we propose to use similar formal analysis methods using common ground structures as the medium within which to both recognize and generate multimodal referring expressions by maximizing the information content provided by each applicable modality.

As common ground structures provide a formal and explainable way of segmenting multimodal content and the information specified by each modal channel, we are also exploring other tasks in which common ground structures may be useful representations.  Some examples include:

\begin{itemize}
\begin{figure}[h!]
    \centering
    \includegraphics[width=.5\textwidth]{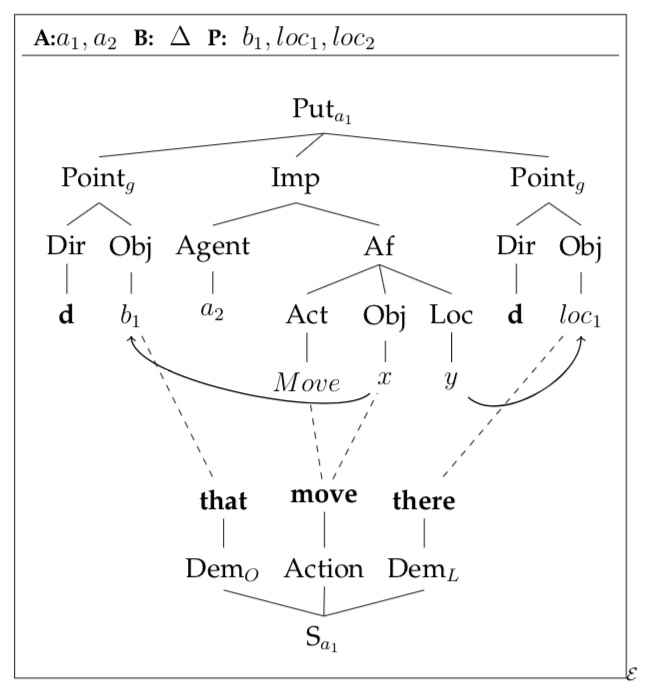}
    \caption{Action command using gesture-language ensemble}
    \label{fig:ensemble}
\end{figure}
    \item {\it Multimodal dialogue parsing}.  Given a situation where both gestures and natural language can indicate both objects and actions or events, common ground structures should be helpful in extracting both object and action information separately from each modality and in disambiguating the information provided by one modality with information from the other (see Fig.~\ref{fig:ensemble}).
\end{itemize}
    
\begin{itemize}
\begin{figure}[h!]
    \centering
    \includegraphics[width=.5\textwidth]{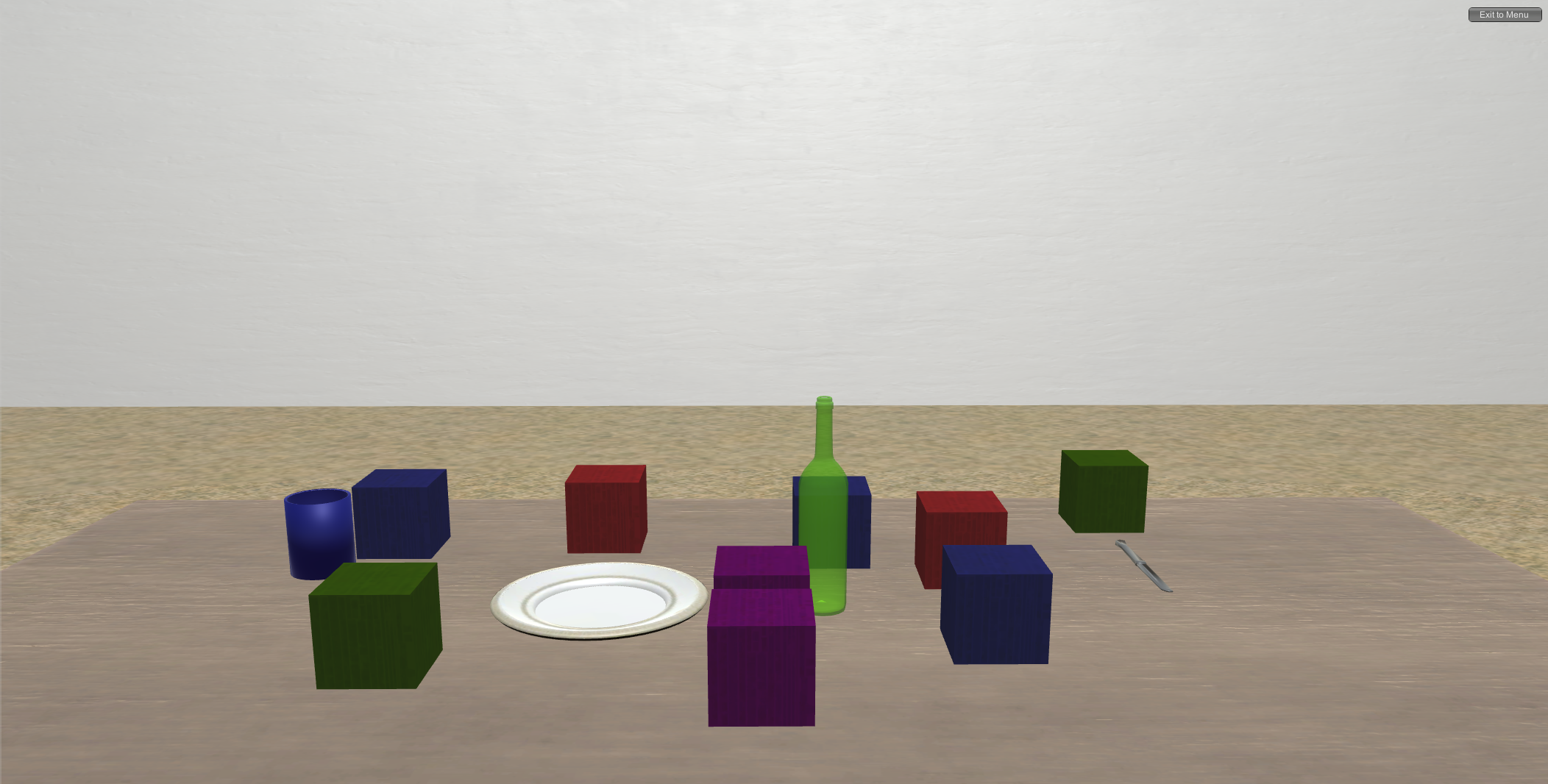}
    \caption{Sample novel situation}
    \label{fig:newscene}
\end{figure}
    \item {\it Scene classification}.  By exploiting the relation sets between objects that populate the belief space, common ground structures can cluster and classify novel scenes and configurations with known examples, providing a way to transfer dialogue or referring strategies from a known situation to a novel one (see Fig.~\ref{fig:newscene}).
\end{itemize}

\begin{itemize}
    \item {\it Intelligent modality switching}.  There may be cases when an agent cannot use one modality or another---e.g., hands are full, prohibiting gesture, or the environment is loud, prohibiting language (cf. Kim et al.  \shortcite{kim2016feature}, Drijvers et al.  \shortcite{drijvers2018hearing})---in this case common ground structures can be deployed to maximize the information content in the remaining available modalities for optimal communication in sub-optimal circumstances.
\end{itemize}

\section*{Acknowledgments}

We would like to thank the reviewers for their helpful comments.  This work was supported by the US Defense Advanced Research Projects Agency (DARPA) and the Army Research Office (ARO) under contract \#W911NF-15-C-0238 at Brandeis University.  The points of view expressed herein are solely those of the authors and do not represent the views of the Department of Defense or the United States Government.  Any errors or omissions are, of course, the responsibility of the authors.

\section{Bibliographical References}
\label{main:ref}

\bibliographystyle{lrec}
\bibliography{References}


\end{document}